\documentclass{article}

\usepackage[nonatbib, preprint]{neurips_2023}

\usepackage[utf8]{inputenc} % allow utf-8 input
\usepackage[T1]{fontenc}    % use 8-bit T1 fonts
\usepackage{hyperref}       % hyperlinks
\usepackage{url}            % simple URL typesetting
\usepackage{booktabs}       % professional-quality tables
\usepackage{amsfonts}       % blackboard math symbols
\usepackage{nicefrac}       % compact symbols for 1/2, etc.
\usepackage{microtype}      % microtypography
\usepackage{xcolor}         % colors
\usepackage{graphicx}
\usepackage[backend=bibtex, style=numeric]{biblatex}
\title{Training neural networks without backpropagation using particles}

\author{%
  Deepak Kumar\\
  \texttt{dipkmr@gmail.com} \\
}

\begin{document}

\maketitle

\begin{abstract}
Neural networks are a group of neurons stacked together in multiple layers to mimic the biological neurons in a human brain. Neural networks have been trained using the backpropagation algorithm based on gradient descent strategy for several decades. Several variants have been developed to improve the backpropagation algorithm. The loss function for the neural network is optimized through backpropagation, but several local minima exist in the manifold of the constructed neural network. We obtain several solutions matching the minima. The gradient descent strategy cannot avoid the problem of local minima and gets stuck in the minima due to the initialization. Particle swarm optimization (PSO) was proposed to select the best local minima among the search space of the loss function. The search space is limited to the instantiated particles in the PSO algorithm, and sometimes it cannot select the best solution. In the proposed approach, we overcome the problem of gradient descent and the limitation of the PSO algorithm by training individual neurons separately, capable of collectively solving the problem as a group of neurons forming a network. Our code and data are available at \href{https://github.com/dipkmr/train-nn-wobp/}{https://github.com/dipkmr/train-nn-wobp/} 
\end{abstract}

\section{Introduction}

We are fascinated by exploring the world of biological neurons in the human brain. Many researchers have spent their whole lives decoding and understanding the workings of the human brain. The researchers conceptualized models to propose the workings of the human brain. One of the early proponents is neural networks. Neural networks are a group of neurons stacked together in multiple layers and modeled to mimic the parts of a human brain. A perceptron was used to classify a set of linearly separable classes through adaptive weight updates \cite{bishop1995book, bishop2006book, duda2001book, gurney1997book}. However, the perceptron was unable to classify the nonlinearly separable data samples. A multiple layer of perceptrons (MLP) was conceptualized to overcome the problem of classifying the nonlinearly separable data samples. The backpropagation algorithm was developed to update the weights from the top (higher) layer to the bottom (lower) layer using a gradient descent approach in the reverse order \cite{rumelhart1986}. Several variants of backpropagation algorithms are developed. A few are Stochastic gradient descent (SGD), RMSProp, Adam optimizer, and Adamax \cite{duchi2011, kingma2015}. One of the difficulties was the vanishing gradient that appeared when the number of layers was higher in a neural network. The activation function was changed from a sigmoid to a rectified linear fashion. The neural node inactive in the forward computation is not updated in the rectified linear activation function. During backpropagation,  it helped to reduce the vanishing gradients problem. The backpropagation algorithms settle to a local minimum while optimizing the loss function constructed for the neural network. However, a few approaches smooth the manifold of the function to improve the optimization and choose a better optimal value.

Particle swarm optimization (PSO) was inspired by the nature of birds flocking and fish schooling \cite{kennedy95particle, kennedy2001book}. It chooses the best neural network from a set of particles (birds). The particles are introduced in the search space. The birds share the value of the function at a particular location with each other. The best value among all locations is considered the current best location for the neural network. The other birds are informed to move closer to the best value while the current best location bird also looks for the better location. The birds update the global best and personal best value to move in a specific direction to reach the present best location. The process is repeated for several ‘N’ times. When the repetition is complete, the global best location is considered the optimal solution for the loss function. The search space is the weights of a neural network, and the location is the value of the weights. The loss function value is computed using the data by passing it through the neural network consisting of the weights of a particular location. The search space on the manifold to be explored is unlimited, so we need to initialize many particles to search larger space, hoping to find the most optimal value for the loss function. However, the manifold of the loss function showed several local minima in a tiny space when the number of layers in the neural network was higher. Due to the problem of several local minima, the neural network training with many layers was difficult, and the results didn’t match the expected outcome.       

We propose a method that doesn’t perform backpropagation for the constructed neural network and follows the principles of particle swarm optimization. The problem of vanishing gradients appearing in the backpropagation is avoided. However, the neural network weights used in the PSO search space were limited due to the local minima within the explored area. So, we split the neural network into individual neuron nodes as a group of particles in the proposed method. Thus, the search space problem of the initial neural network is divided into many sub-spaces. With this approach, it was easier for the particles to update the weights of single neuron nodes.

The main contributions of the proposed method are as follows:
\begin{itemize}
\item Neural networks are trained without using backpropagation
\item Each neuron node is considered a sub-problem, and particle swarms are deployed for each neuron node
\item The weights of each neuron node are updated separately
\item The collective effort of all neurons learn the complex pattern in the dataset, which showcases that the solution is a collection of small agents to form bigger agents
\end{itemize}

\section{Related Work}

Several approaches have been explored to train neural networks, improving backpropagation or without it. A reward penalty function was introduced to train multiple layer perceptron or neural networks without backpropagation \cite{andrew1987}. The global performance of the network is used as an estimate for the gradient instead of an exact gradient in \cite{andrew1987}. We need forward and backward passes in backpropagation during training, and the weights are to be transported. The weights of neural networks are updated without transporting the weight information, but changes in the weight values are transported in \cite{kolen1994}. A refined version of backpropagation was proposed to work on the network weights. An implicit error feedback is used in place of backpropagation \cite{brandt1996}. Another kind of refinement was proposed on backpropagation in terms of linearity. A linear backpropagation approach is replaced by a non-linear backpropagation as part of refinement \cite{hertz1997}. 

The helpfulness of forward matrix computation in training the neural network was explored in \cite{wilamowski2010}. A set of random search options to minimize the convex functions was proposed and showed that the backpropagation still holds good as a rule to move forward or accelerate the action \cite{nesterov2017}. The error propagation with random synaptic weights helps to learn, which approximately aligns with their feedforward synaptic weights. The feedback alignment approach demonstrates similarity with the backpropagation approach \cite{lillicrap2014, lillicrap2016}. Random feedback will help backpropagation, and the constraints are not required. An exploration of mitigating the exploding and vanishing gradients is also covered \cite{lillicrap2014, lillicrap2016}. A method is applied to RNN by generating synthetic gradients while performing forward passes. The updates between each step are decoupled in this method to show the decoupled networks can learn independently in \cite{jaderberg2016}. The system in \cite{haber2018} uses Ensemble Kalman Filter particles to update the weights. However, the condition is strict in improving the objective function. A machine learning problem is posed as an inverse problem, and an ensemble Kalman filter is used for non-derivative objective functions \cite{kovachki2019}.

The multiple layer networks as autoencoders at each stage to build the representation in the system lead to the proposal of the Neural gradient representation (NGRAD) framework to demonstrate the core principles of backpropagation \cite{lillicrap2020}. Hilbert--Schmidt independence criterion (HSIC) is used to train deep learning. This approach helps mitigate exploding and vanishing gradients \cite{ma2020}. An automated derivative computation for the functions to learn in a single forward run was proposed in \cite{baydin2022}. It is an unbiased estimate of the gradients. Wide neural networks are trained using the neural tangent kernel (NTK) regime, equivalent to kernel regression \cite{boopathy2021}. A unified likelihood ratio to gradient estimation in forward propagation is proposed to promote likelihood ratio better than backpropagation \cite{jiang2023}. Two passes are performed while using backpropagation to train a network. One is a forward pass, and the other is a backward pass. The forward pass replaces the backward pass, and there are two forward passes in \cite{hinton2022}.

The evolutionary algorithms are used in deep learning to obtain better architectures. Generally, the genetic algorithm approach is followed to improve the solution iteratively \cite{felipe2017, valdez2019, edgar2021}. Sometimes, gradients are also used to boost the process \cite{shangshang2022}. One concept was to evolve the architecture topology by introducing evolutionary algorithms \cite{gregory2016}. All evolutionary algorithms avoid getting stuck in local optimal solutions and explore the parameter space for better solutions. It may look advantageous but comes with a computational cost to obtain a better solution. In natural evolution strategies (NES), a correlation matrix is formed to generate new children, and then the best among the children is used to improve the solution \cite{daan2014}. An exponential term was added to speed up the matrix computation \cite{tobias2010}. An efficient method was developed, introducing importance sampling \cite{yi2009}. A parallel implementation was proposed to demonstrate the scalability of NES \cite{salimans2017}. In all these approaches, the algorithms consider the entire parameter or hyperparameter space for the solutions. However, we propose a solution that considers each neuron as an independent entity.

\section{Particle Swarm Optimization (PSO)}

Particle swarm optimization (PSO) was proposed for optimizing the weights of a neural network \cite{kennedy95particle}. Literature on all derivative-free optimization algorithms applied to non-derivative functions with several convex and non-convex problems in \cite{rios2013}. PSO is applied to numerous applications for optimizing non-linear functions \cite{kennedy2001book}. PSO evolved by simulating bird flocking and fish schooling. The advantages of PSO are conceptually simple and easy to implement. Particles are deployed in the search space, and each particle is evaluated against an optimization function. The best particle is chosen as a directing agent for the rest. The velocity of each particle is controlled by both the particle’s personal and global best. During the movement of the particles, a few of them may reach the global best. Several iterations are required to attain the global best.

Let $X = {x_1, x_2, ..., x_k}$ be the particles deployed in the search space of the optimization function, where $k$ is the number of particles and $V = {v_1, v_2, ..., v_k}$ be the velocities of the respective particles. $x_i, v_i \in R^n$ for all the $k$ particles. A simple PSO update is as follows.

Velocity update equation
\begin{equation}
v_i^j = wv_i^{j-1} + c_1 r_1 (x_{bi} - x_i^{j-1}) + c_2 r_2 (x_{bg} - x_i^{j-1}),
\end{equation}

where $w$ is the weight for the previous velocity; $c_1 , c_2$ are constants and $r_1 , r_2$ are random values varied in each iteration. $x_{bi}$ is the personal best value for particle $i$ and $x_{bg}$ is the global best value among all the particles. $v_i^j$ is the updated velocity of the $i^{th}$ particle in the $j^{th}$ iteration and $v_i^{j-1}$ is its velocity in the $(j-1)^{th}$ iteration, $x_i^{j-1}$ is the position of the $i^{th}$ particle after $(j-1)^{th}$ iteration.

The updated velocity is added to the existing position of the particle. The position update equation is
\begin{equation}
x_i^j = x_i^{j-1} + v_i^j
\end{equation}

where $x_i^j$ is the updated position of the $i^{th}$ particle in the $j^{th}$ iteration.

\subsection{Craziness term}
The original PSO paper had removed the craziness term in the velocity update equation because the algorithm looked realistic and sufficient. When PSO was applied in \cite{kumar2014} for optimization of convex optimization functions, the particles used to get stuck in particle direction due to the boundary conditions of the convex functions. This is particularly relevant when there is a limited set of directions in the particles to move forward \cite{kumar2016}. However, this is not the case when the number of particles is large. The relevance of the craziness term is minimized with the number of particles. Then, \cite{kumar2016} introduced the craziness term in the velocity update equation to use less number of particles. With the introduction of the craziness term, the particles started exploring the search space and reached the optimal value in convex optimization functions and non-convex functions with boundary conditions.

\begin{equation} 
v_i^j = wv_i^{j-1} + c_1 r_1 (x_{bi} - x_i^{j-1}) + c_2 r_2 (x_{bg} - x_i^{j-1}) + c_3 r_3
\end{equation}

where $c_3$ is a constant and $r_3 \in R^n$ is a random vector similar to a particle position.

\section{Neural networks (NN)}

\begin{figure*}[!t]
\centering
\includegraphics[scale=0.75]{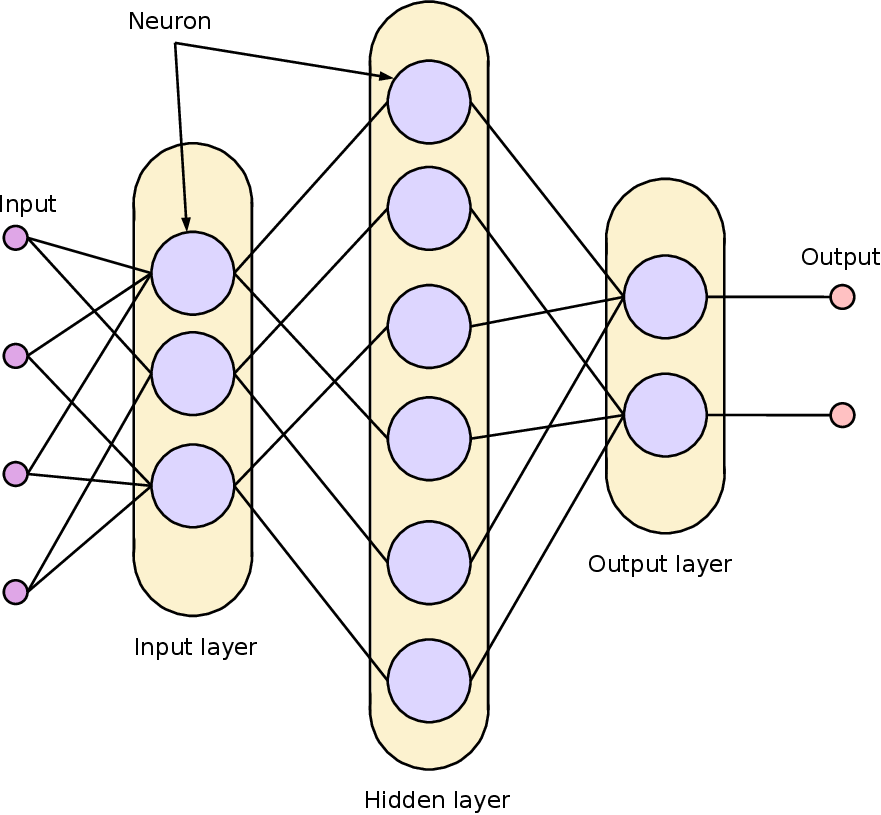}
\caption{A three layer (MLP) neural network consisting of input, hidden, and output layers. There are 3, 6, and 2 neurons in input, hidden, and output layers, respectively for illustration purpose.}
\label{fig:figure1}
\end{figure*}

Neural networks are constructed by placing many neurons in multiple layers. Multi-layer perceptron (MLP) is a general way of understanding neural networks. It consists of three layers: the input layer, the hidden layer, and the output layer. There may be one or more hidden layers in a neural network. A simple MLP is shown in Figure \ref{fig:figure1} with the three layers. The backpropagation is used to train neural networks. We compute the loss function $f(x)$ for a given data during training (which is known as forward pass) and use it as a reference to update the weights of the network (which is known as backward pass) through a variety of gradients.

\subsection{Individual neuron node}

The loss function is tightly dependent on the combination of all nodes in a network. If the weights of a neuron change in a node, then it is propagated to all higher-layer nodes in the network. We would like to know the contribution of each node in the network. We can perform this operation by fixing the weights of all nodes except the considered node. We can change the considered node weights and compute the loss function for the network. We observe similar behavior or patterns of local minima in the loss function. Instead of a complete network, we approach individual neurons to find the local or optimal value.  

\begin{figure*}[!t]
\centering
\includegraphics[scale=0.75]{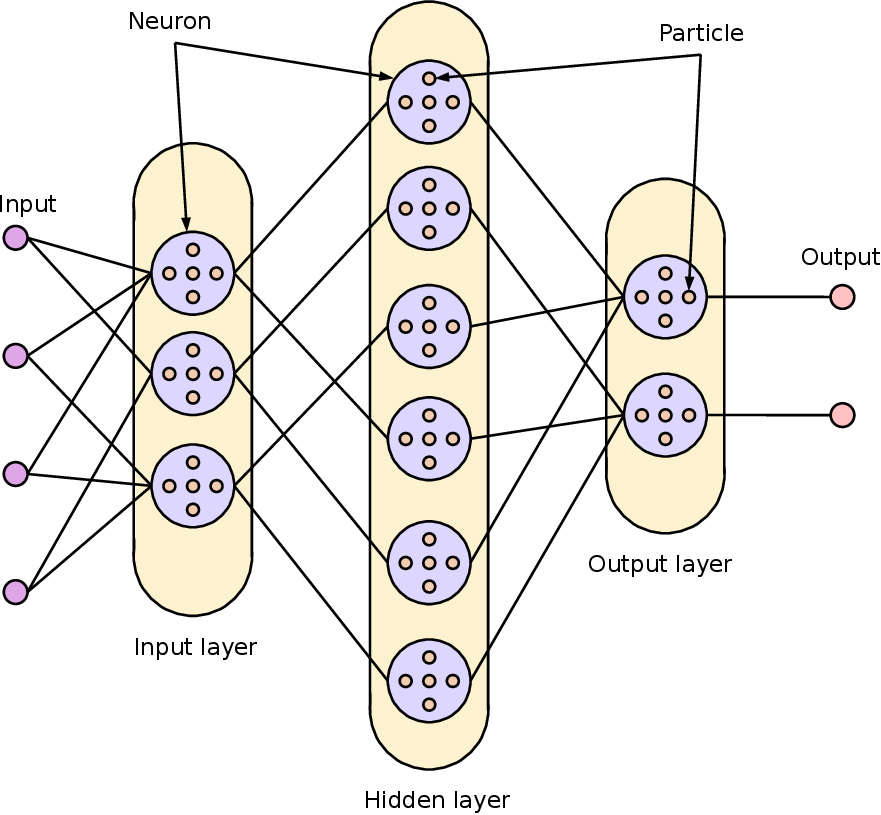}
\caption{A three layer neural network with each node showing `k' particles. The `k' particles update their weights using PSO approach. The number of particles is 5 for illustration purpose in the network.}
\label{fig:figure2}
\end{figure*}

In a neural network, we shall isolate a single neuron node. Here, we fix the weights of all other neurons except the isolated node. The weights of the isolated nodes are perturbed or changed randomly based on the principles of PSO. We perform weight changes for `k' times to match the particle swarm optimization (PSO) approach. In Figure \ref{fig:figure2}, we have placed `k' particles in each neuron node. Now, we have `k' different weights for the isolated node. We compute the loss function value by plugging in the `k' different weights of the isolated node. We get `k' different loss values. We shall choose the minimum value and its weight as the better performance of the node. The contribution of other nodes appears in the computation of the loss function and does not involve in the isolated node weight updates. The `k' different loss values allow us to select the minimum loss value without any other information from the other nodes. The differentiation required for backpropagation is avoided through the PSO approach. Here, we can choose the direction of differentiation as an improvement factor for the weight updates in the isolated node as an alternative approach. We can independently and parallelly perform weight update operations for all the nodes. In each iteration, all the nodes will choose their minimum loss function value as the possible weight.

The single loss function of a neural network is split into multiple loss functions attached to each node. The loss functions can be expressed $f_{ik}$ for each node, where $i$ is the $i^{th}$ node in the layer and $k$ is the $k^{th}$ particle of the node. We select the best weight through the minimization of the loss functions.

\begin{equation}
f_{i} = \min_{w_{ik}} f_{ik} 
\end{equation}

where $w_{ik}$ is the weight of the $k^{th}$ particle in the $i^{th}$ node.

The major problem with this approach is the weights chosen by some nodes are better, and for others, it may be worse for a given data. Since the particles keep hopping from one position to another position. It is difficult to keep track of the best weights for the nodes. The nodes together perform runaway operations for unseen data, causing difficulty in placing a check on the nodes. Due to this, stability doesn’t exist with the weight updates performed without placing measures to counter the runaway problem faced by the network.

\begin{figure*}[!t]
\centering
\includegraphics[scale=0.68]{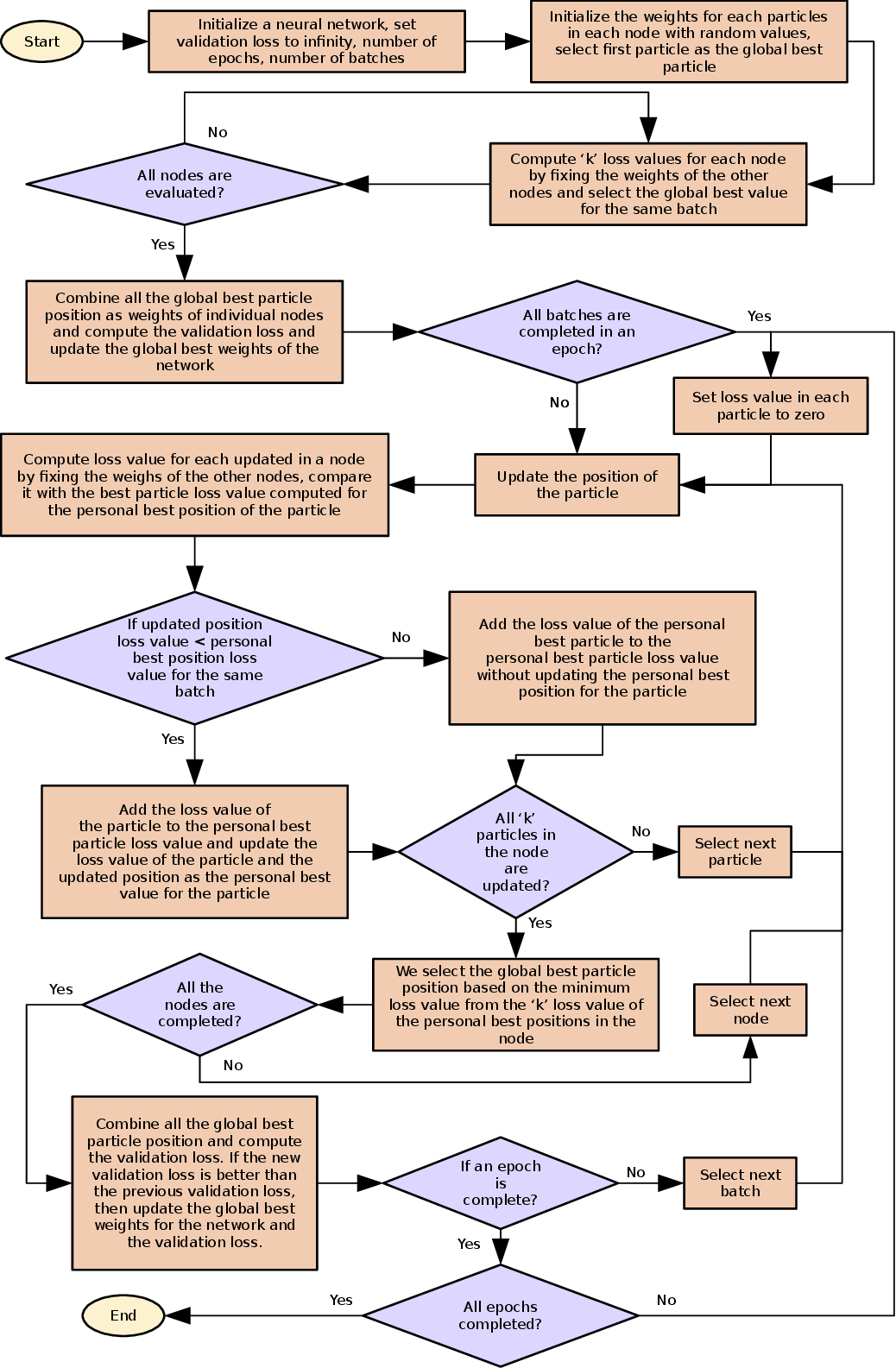}
\caption{The flowchart of the proposed method including a batchwise style of training the network.}
\label{fig:figure2f}
\end{figure*}

\section{Proposed method}

The individual nodes act independently by fixing the weights of all other nodes. The control of the individual nodes is difficult. When all the nodes are updated, we select the best weights for each node and combine them. All the node weights combination forms the best network weights for the problem. Now, we compute the training/validation loss for the data. After the weight update, we will come to know whether the loss has decreased or not. If the loss has been reduced, the new weights are the best weights for the network, or the previous best weights are used as the best weights. If the best weights are rewired back to each node as the global best for each node, then the network sticks to the original best weights and restricts the exploration of the manifold by the nodes. The isolated nodes are unrestricted by any performance control metric. However, the group of all the best weights from all the nodes is evaluated to control the network behavior. The performance measure condition used to select the best weights controls the behavior of the nodes in a grouped form and forces the nodes to reach the best optimal position. The performance measures may be training loss, validation loss, training accuracy, or validation accuracy. Independent/validation data different from the training data exerts more control on the weight update for the network. Figure \ref{fig:figure2f} shows the flowchart of the proposed method in a batch-wise training approach.

In batch-wise training, we split training data into several batches and passed them to the network. Each batch of data is new and unseen for the network during training. When PSO particles are updated in the nodes for one batch, the personal and global best are highly likely to change after the update due to the movement of the particles to a new position. The individual best position of each particle is retained and evaluated for the next batch. If the personal best position is best for the new batch, then the personal best position is retained else it is updated with the new position found for the new batch. The concept of the global optimal position is unknown to the node, and the node reaches the global optimal position after several iterations. Now, the PSO particles change their position randomly through some updates. However, the node has already reached the global optimal position, and the updates would have changed the position of the particles, which is not optimal. Since the particles are not restricted, the particles move out of the global optimal position. To check the global optimal position, we compare the loss function value for the next batch with the global optimal position and the updated position of the particles. If the updated position of the particles has a better loss, then the particle position is updated else the previous optimal position is retained without moving to a new position. In this way, the movement of particles after reaching the global optimal position is controlled. 

When we split the training data into batches, the loss function $f_{ik}$ is divided into batches. Initially, the loss function is set to zero after each epoch. The particles are updated after each batch. So, we have the previous best position and the newly updated position for the next/new batch. We compute the loss function for both the particle positions and compare their loss functions. The final loss function is updated as shown in the equation.  

\begin{equation}
f_{ik}^{j}=\left\{\begin{array}{ll}
f_{ik}^{j-1} + f_{ik}(w_{ik}^{j}), &  f_{ik}(w_{ik}^{j}) < f_{ik}(w_{bik}^{j-1}). \\ 
f_{ik}^{j-1} + f_{ik}(w_{bik}^{j-1}), & \mbox{otherwise}.
\end{array}\right.
\end{equation}

\begin{table*}[ht!]
  \begin{center}
%    \caption{The steps followed in the proposed method to train a neural network without backpropagation}
    \label{tab:table1}
    \begin{tabular}{l} % <-- Alignments: 1st column left, 2nd middle and 3rd right, with vertical lines in between
   \hline
   \hline
   \textbf{Algorithm 1:} The steps followed in the proposed method to train a neural network\\
   without backpropagation using particles.\\
   \hline
\textbf{Input:} Training data and labels, independent validation data and labels\\
(may be same as training data if needed)\\
\textbf{Output:} The best weights of the network after training the network\\
\textbf{Network:} Initialize network with required number of layers and number of neurons\\
\hline
\textbf{1:} Initialize the network with the nodes in each input, hidden, and output layer.\\
Validation loss is set to Infinity. The number of epochs. The number of batches.\\
\textbf{2:} Initialize the weights for each particles, which are personal best for the particle,\\
in each node with random values. \\ 
\textbf{3:} Select the first particle as the global best particle in each node.\\
\textbf{4:} Compute `k' loss values for each node while fixing the weights of the other nodes\\
and select the global best value for a batch.\\
\textbf{5:} Repeat step 4 for all the nodes.\\
\textbf{6:} Combine all the global best particle position as wieghts of individual nodes and\\
compute the validation loss and update the global best weights of the network.\\
\textbf{7:} if all batches are completed in an epoch, then set loss values in each particle to zero\\
\textbf{8:} update the position of each particle\\
\textbf{9:} compute loss values for each updated particle in a node while fixing\\
 the global best weights of the other nodes and also compare it with the loss value\\
 computed for the personal best position of that particle.\\
\textbf{10:} \hspace{2mm} if the loss value of the updated particle position is better than the \\
loss value of the personal best particle position for the batch, then add the loss value of \\
the particle to the personal best particle loss value and update the loss value of the\\
 particle and the updated position as the personal best value for the particle.\\
\textbf{11:} \hspace{2mm} else, add the loss value of the personal best particle to the\\
personal best particle loss value without updating the personal best position for the particle. \\ 
\textbf{12:} After repeating the steps 8, 9, 10, and 11 for all `k' particles in the node. We select the\\
global best particle position based on the minimum loss value from the `k' loss value of \\ the 
personal best positions. \\
\textbf{13:} if all nodes are not completed, then repeat the steps 8, 9, 10, 11 and 12 for all\\
 the nodes, else go to next step.\\
\textbf{14:} Combine all the global best particle position and compute the validation loss. \\
If the new validation loss is better than the previous validation loss, then update \\
the global best weights for the network and the validation loss.\\
\textbf{15:} If an epoch is complete, then go to step 7, else go to step 8.\\
\textbf{16:} Repeat steps 7 - 15 for all the batches and epochs.\\
      \hline
      \hline
    \end{tabular}
  \end{center}
\end{table*}

where $w_{ik}^{j}$ is the weight of the $k^{th}$ particle in the $i^{th}$ node for the $j^{th}$ batch, $w_{bik}^{j-1}$ is the personal best weight of the $k^{th}$ particle in the $i^{th}$ node for the ${j-1}^{th}$ batch, $f_{ik}^{j-1}$ is the loss for the ${j-1}^{th}$ batch, and $f_{ik}^{j}$ is the loss for the ${j}^{th}$ batch .

The proposed method algorithm is presented on the next page. We initialize a neural network with nodes in each layer. Each node in the network gets its particle to explore the search space. There may be ‘k’ particles in each node that act independently. The random weights are assigned for each particle. In the first batch or epoch, the first particle weight is chosen as the global best weight of a neuron to perform the computation before selecting the best particle position. The training data is split into batches. We set the loss function in each node to zero and the validation loss function to Infinity. The loss function value is set to zero after every epoch. The particle position, the node weight, is used to compute the loss value for the given batch while fixing all the other node weights in the network. We obtain ‘k’ loss values and select one of them as the global best value for the node. All the nodes are updated similarly, and the global best values are computed for the entire network. Now, we compute the training/validation loss of the independent set to check whether the loss is better and update the global best value and the global best weights. From the next batch, we compute two losses: a loss value for the updated position and a loss value for the personal best position of the particle. We compare both loss values and update the loss and position values depending on the loss values. The comparison is performed when the data is used batch-wise. We do not use comparison when the complete data is utilized in the loss value computation. The comparison provides an overlap between the data batches for a position. The process is repeated for all the batches. The process is repeated for ‘n’ epochs, or the loss is not updated for certain epochs.

\section{Experiments and Results}

We have created synthetic examples to simulate a real dataset and demonstrate the functioning of the proposed method. The complexity of the examples increases with the number of classes and the addition of nonlinearity in the examples. We have used our proposed method on the two real datasets. All the datasets are normalized to zero mean and unit variance except the linearly separable classes dataset. We have also compared the proposed method with the other MLP implementations. The number of epochs is set to 20 when the neural networks are trained for simulated datasets.

\begin{figure*}[!ht]
\centering
\includegraphics[scale=0.341]{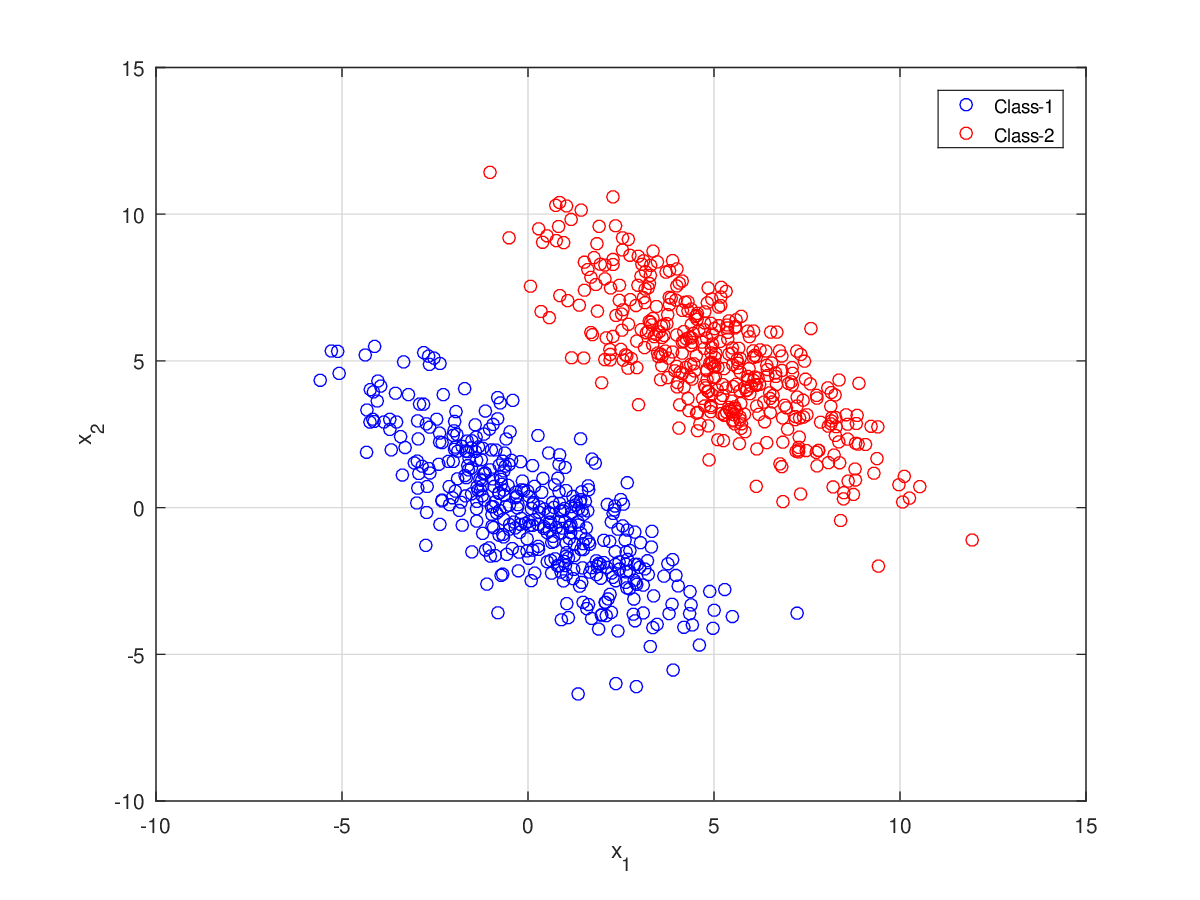}
\includegraphics[scale=0.341]{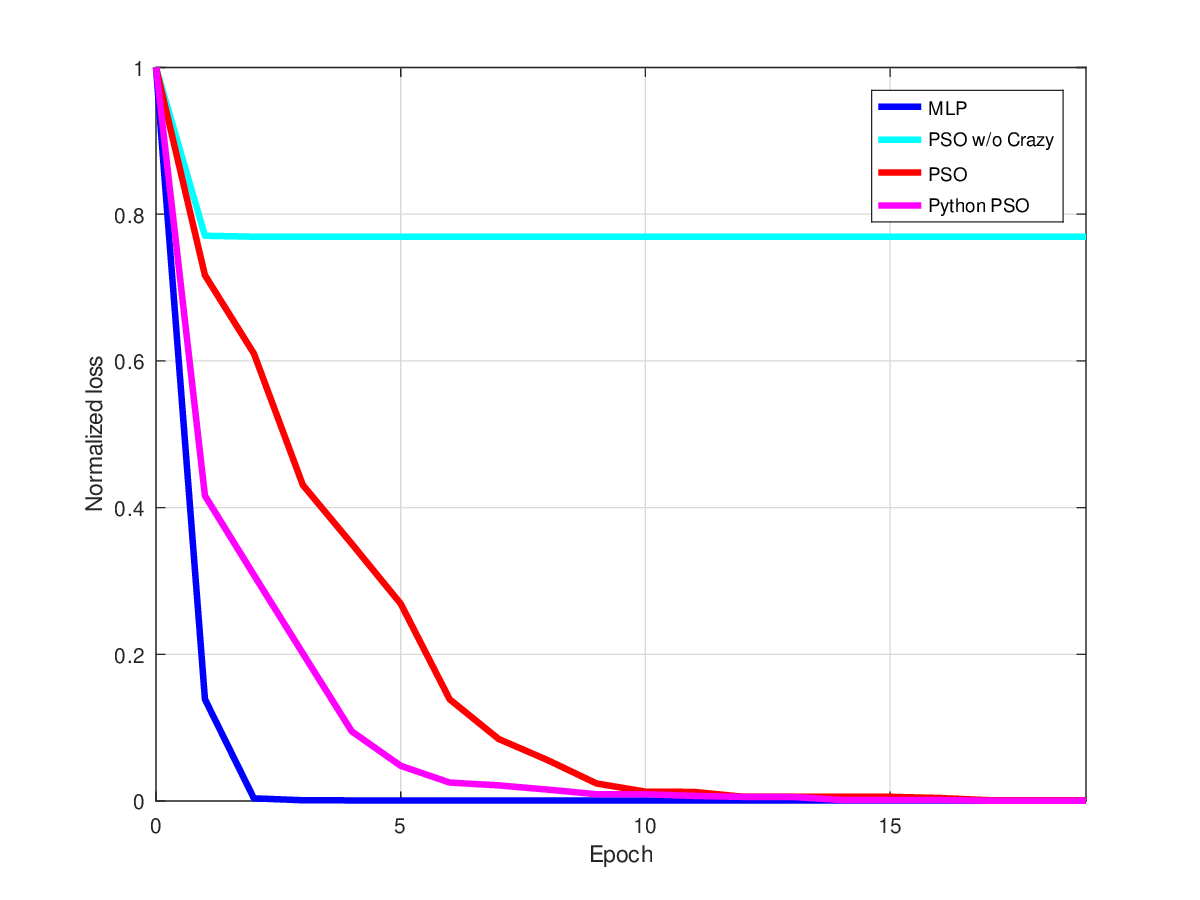}
\caption{The synthetic data with linearly separable classes. The normalized loss function after each epoch is shown for basic MLP architecture, PSO without craziness term, the proposed method using Octave and Python.}
\label{fig:figure3}
\end{figure*}

\begin{figure*}[ht!]
\centering
\includegraphics[scale=0.341]{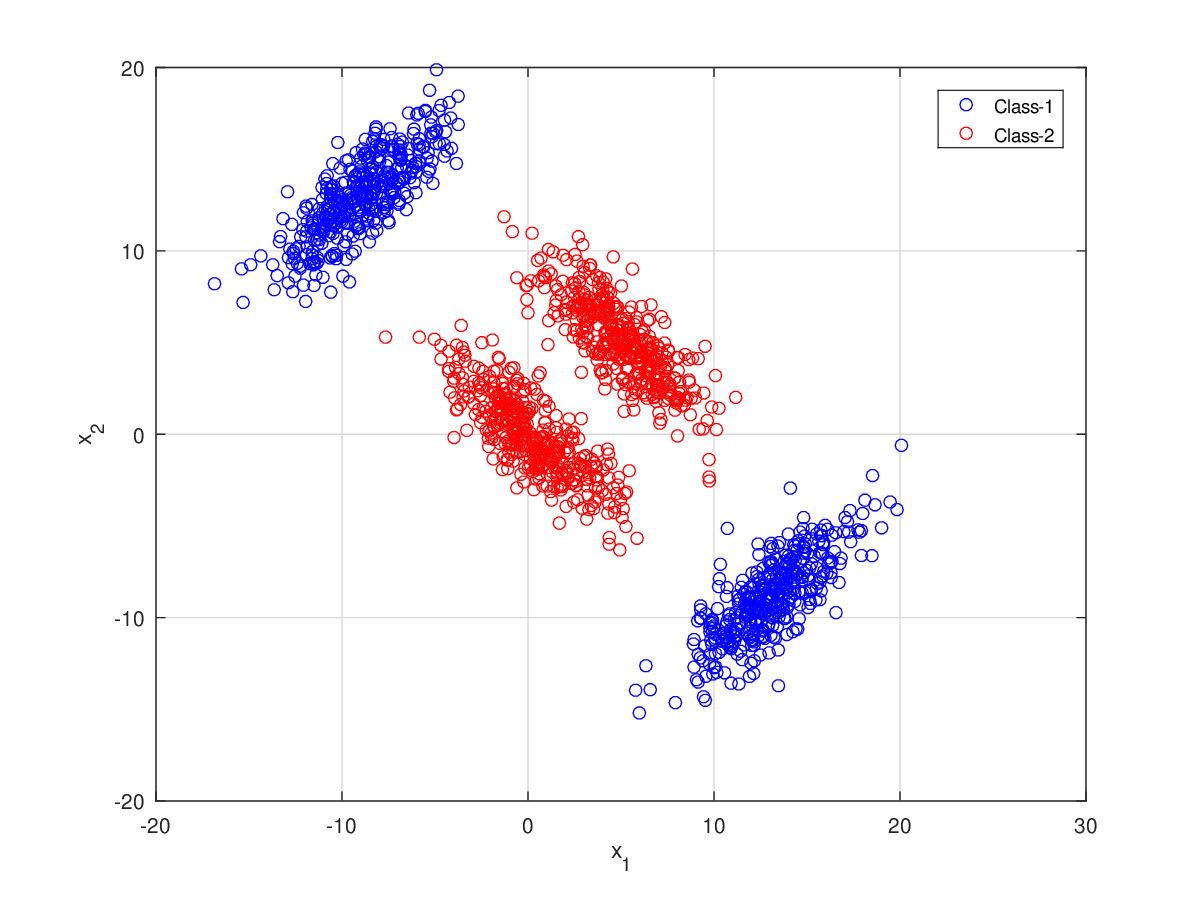}
\includegraphics[scale=0.341]{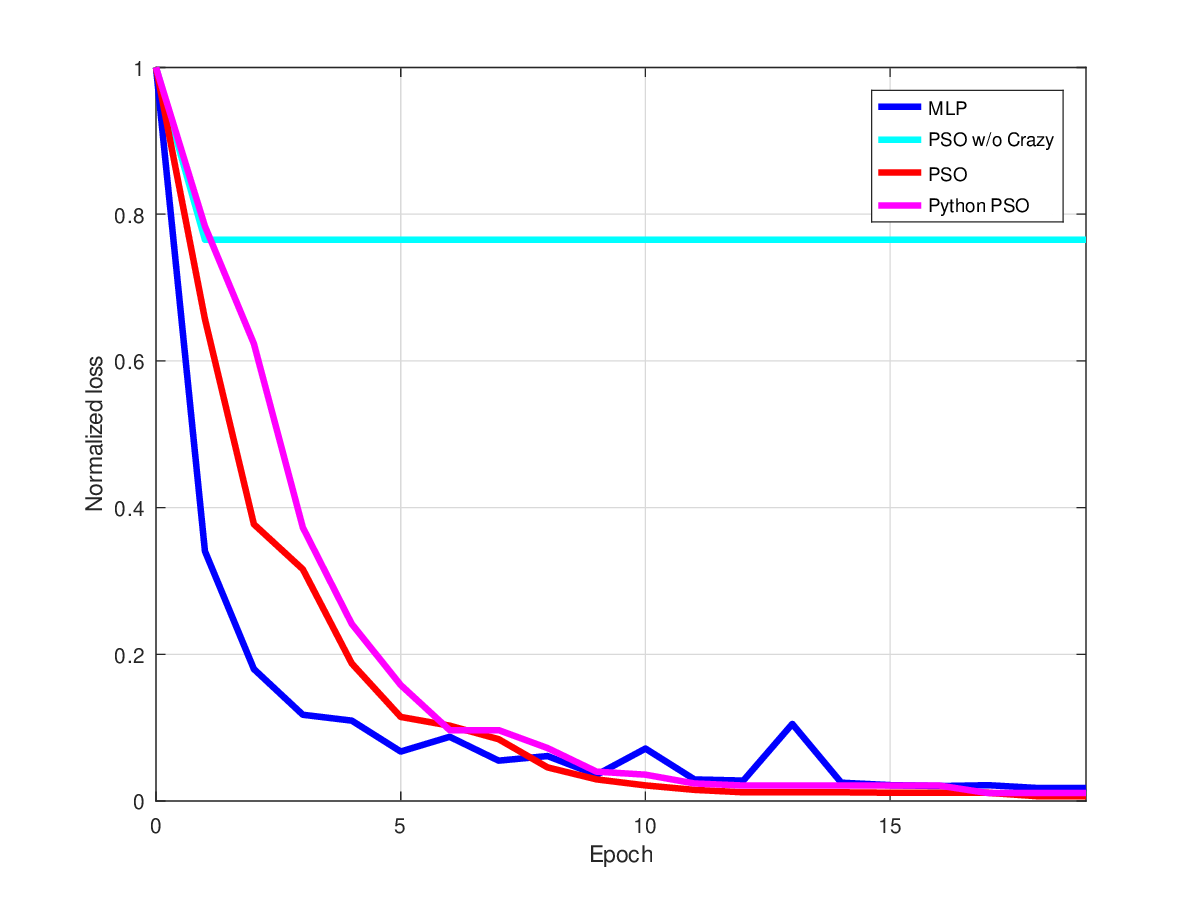}
\caption{The synthetic data with nonlinearly separable classes. The normalized loss function after each epoch is shown for basic MLP architecture, PSO without craziness term, the proposed method using Octave and Python.}
\label{fig:figure4}
\end{figure*}

\subsection{Linearly separable classes}

Two classes are generated in this synthetic example using Gaussian mixtures, as shown in Figure \ref{fig:figure3}. A single perceptron with mean square error (MSE) loss is used as the MLP implementation. The implemented MLP is based on the Pytorch framework with a stochastic gradient descent (SGD) algorithm, which quickly finds the minimal solution. A neural network with 4 nodes in the input layer and 2 in the output layer is constructed to classify the samples. A binary cross-entropy (BCE) loss function is used as the loss function to compute the loss in the PSO-based solutions. The PSO-based solution without the craziness term gets stuck in local minima and doesn't go further to provide a better solution. However, the proposed method using Octave and Python programming languages follows the gradient descent and reaches a minimal solution.
 
\subsection{Nonlinearly separable classes}

There are two classes based on four Gaussian mixtures in this synthetic example, and they are nonseparable, as shown in Figure \ref{fig:figure4}, which cannot be solved using a single perceptron. An MLP architecture is needed to solve this generated synthetic data. A neural network with 4 nodes in the input layer and 2 in the output layer is constructed to classify the samples. The MLP implementation has one node as the output node and is trained using MSE loss, whereas the PSO implementation has two nodes in the output layer and is trained using BCE loss. The PSO solution without craziness terms gets stuck in local minima. The MLP using SGD for backpropagation takes extra time to reach a minimal solution. The proposed method using Octave and Python programming languages reaches a minimal solution and follows the gradient descent approach.

\begin{figure*}[!ht]
\centering
\includegraphics[scale=0.341]{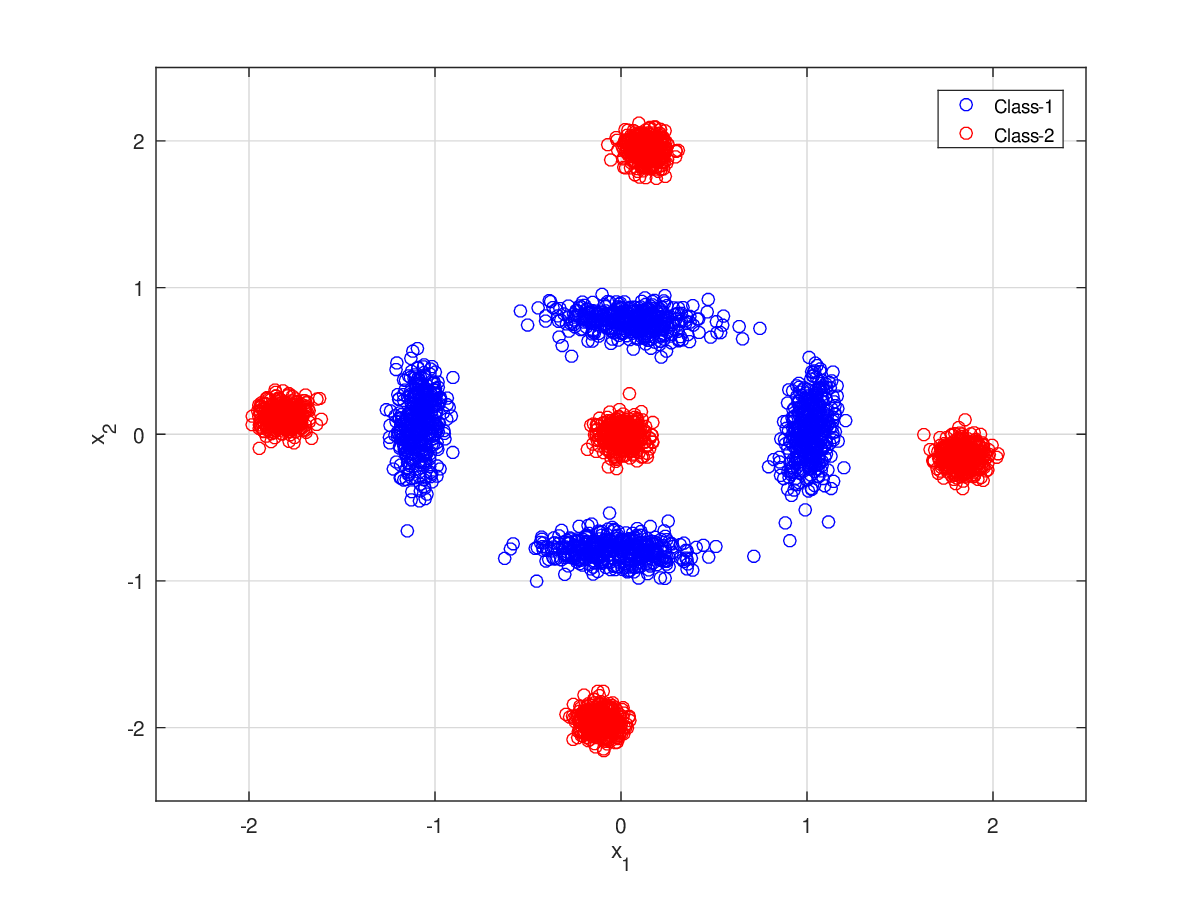}
\includegraphics[scale=0.341]{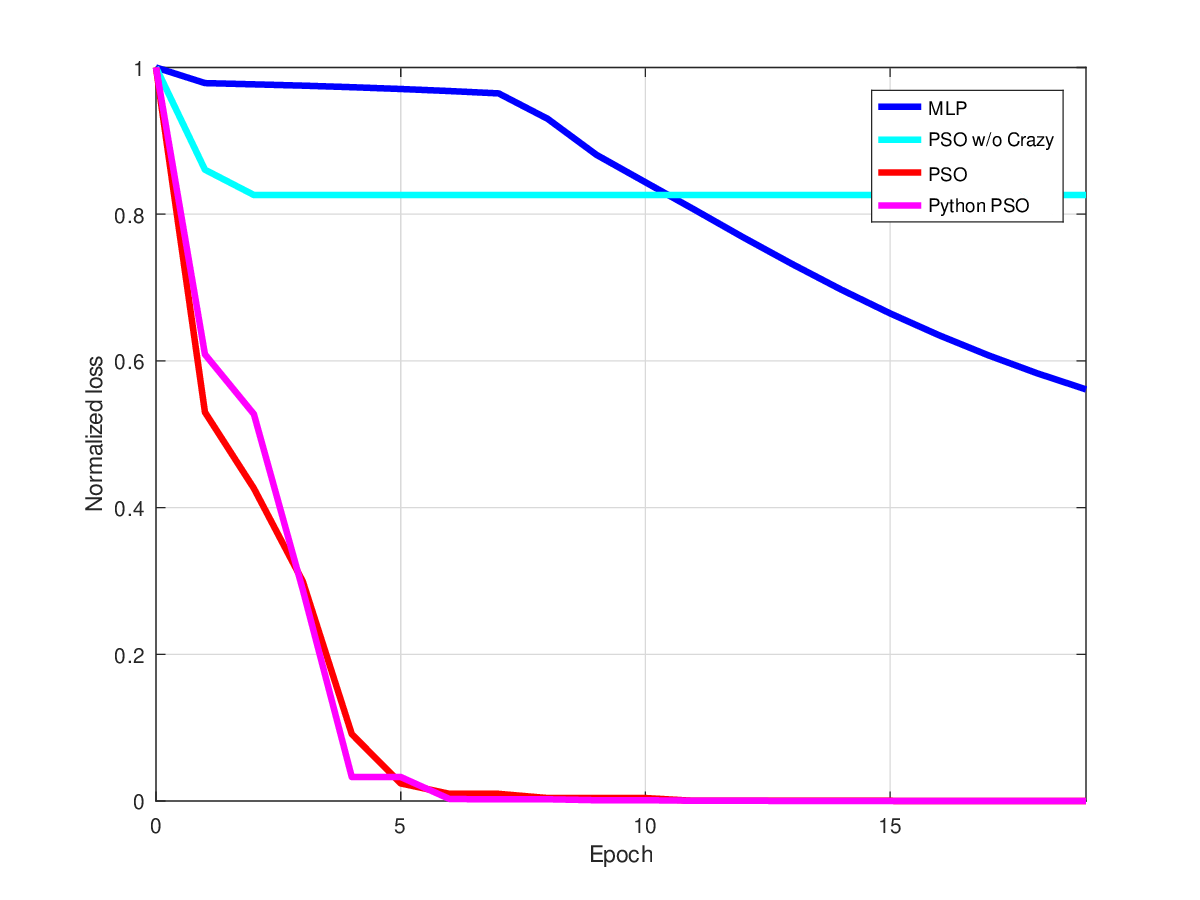}
\caption{The synthetic data with increased nonlinearity. The normalized loss function after each epoch is shown for basic MLP architecture, PSO without craziness term, the proposed method using Octave and Python.}
\label{fig:figure5}
\end{figure*}

\begin{figure*}[!ht]
\centering
\includegraphics[scale=0.341]{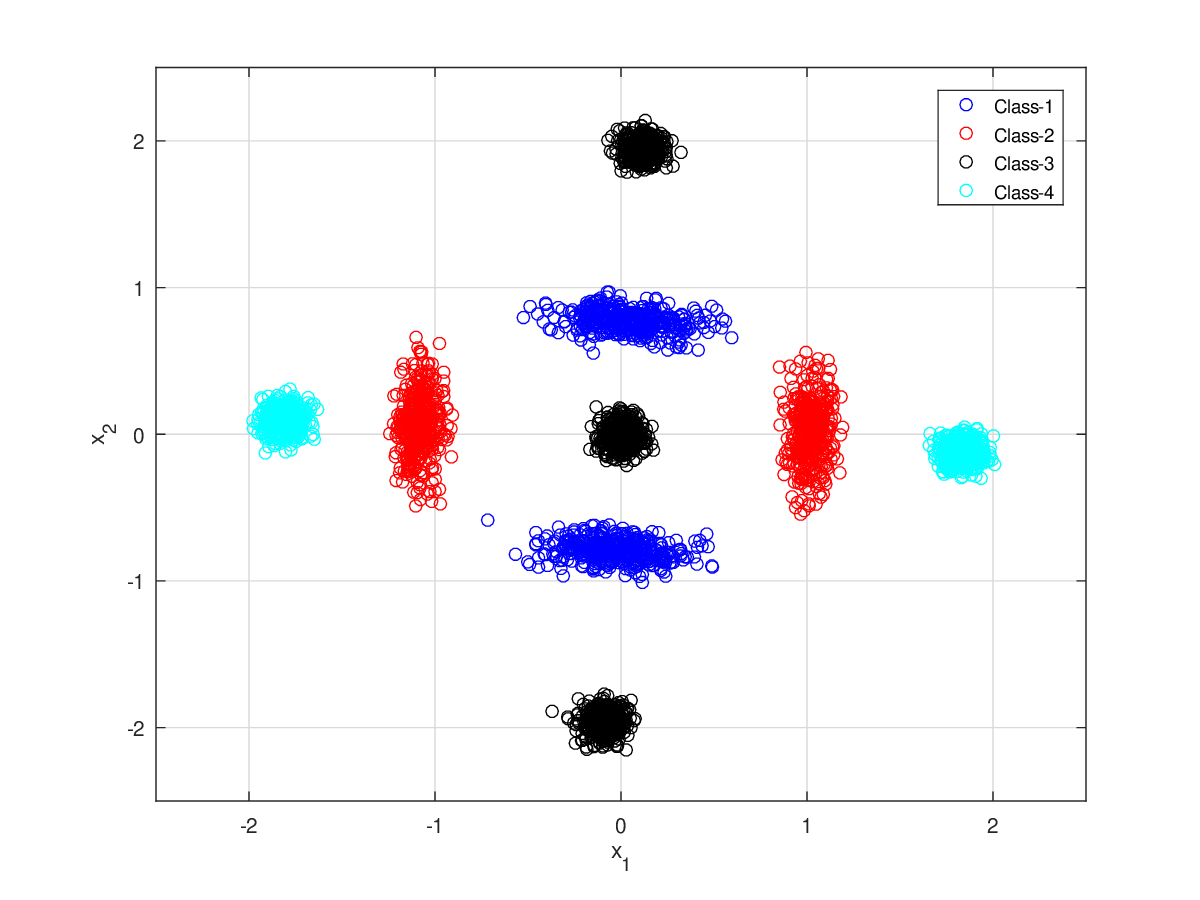}
\includegraphics[scale=0.341]{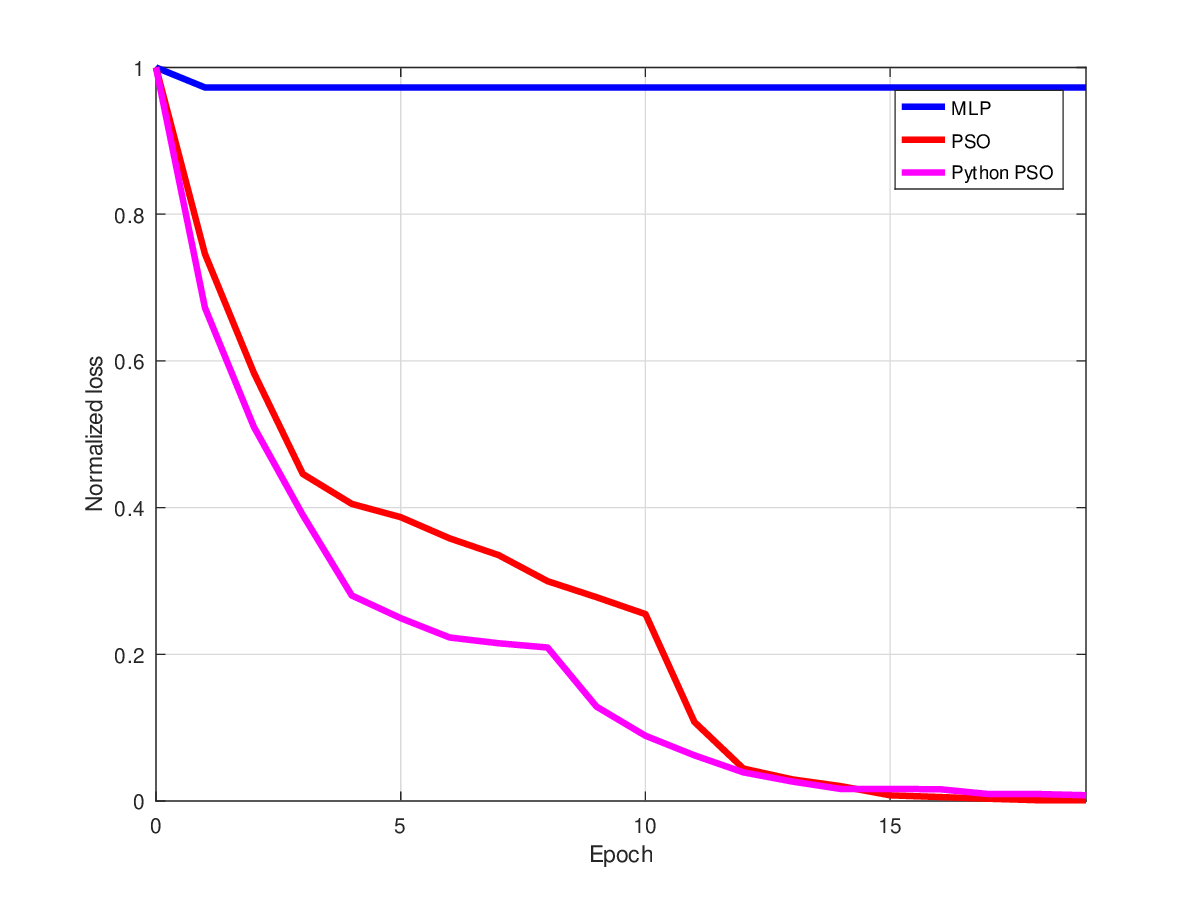}
\caption{The synthetic data with increased number of classes. The normalized loss function after each epoch is shown for basic MLP architecture, the proposed method using Octave and Python.}
\label{fig:figure6}
\end{figure*}

\subsection{Increased nonlinearity}

There are two nonseparable classes based on nine Gaussian mixtures in this synthetic example, as shown in Figure \ref{fig:figure5}. A two-layer neural network faces difficulty in solving the problem. A three-layer network or a higher number of layers solves the problem. All the approaches use 8 nodes as the input layer, 6 as the hidden layer, and 2 as the output layer. A binary cross-entropy (BCE) loss function is used as the loss function to compute the loss in the solutions. The PSO-based solution without the craziness term gets stuck in local minima. However, the proposed method using Octave and Python programming languages follows the gradient descent and reaches a minimal solution. We can observe that the gradient descent-based backpropagation takes time to get to a minimal solution.
 
\subsection{Increased number of classes}

There are four nonseparable classes based on nine Gaussian mixtures in this synthetic example, as shown in Figure \ref{fig:figure6}. A three-layer network or a higher number of layers solves the problem. All approaches use 8 nodes as the input layer, 6 as the hidden layer, and 4 as the output layer. A cross-entropy (CE) loss function is used as the loss function to compute the loss in the solutions. The MLP  with backpropagation gets stuck in a local minima many times, as shown in Figure \ref{fig:figure6}. However, the proposed method using Octave and Python programming languages follows the gradient descent and reaches a minimal solution. The Gaussian mixture in the middle section of the data is not correctly classified several times because of strong local minima in the solution. The particles manage to move out quickly and reach a better solution.

\subsection{Rice dataset}

A set of 3810 rice grain images were collected from the two species (Cammeo and Osmancik) \cite{cinar2019}. There are seven morphological features extracted from each image. The morphological features are area, perimeter, major axis length, minor axis length, eccentricity, convex area, and extent. We perform 4-fold cross-validation and 10 trials. We construct a two-layer neural network with 5 nodes in the input layer and 2 in the output layer. We use cross entropy (CE) loss to compute the loss and train the neural network. The number of epochs is set to 100. Table \ref{table_1} shows the tabulated accuracies of the proposed method with different machine learning methods. We observe that the values of accuracy, specificity, and F1-score fall within a sigma range of the reported results in Table \ref{table_1}. 

\subsection{Dry bean dataset}

A set of 13611 dry bean images were collected from 7 different dry beans (Seker, Barbunya, Bombay, Cali, Dermosan, Horoz, and Sira) \cite{koklu2020}. There are 16 features consisting of 12-dimensional features and 4 shape-form features. We perform 13-fold cross-validation on this dataset. We construct a four-layer neural network with 16 nodes in the input layer, 12 in the first hidden layer, 3 in the second hidden layer, and 7 in the output layer, as mentioned in \cite{koklu2020}. We use cross entropy (CE) loss to compute the loss and train the neural network. The number of epochs is set to 100. Table \ref{table_2} shows the tabulated accuracies of the proposed method with different machine learning methods. All the performance metrics are close to each other in the last three rows of Table \ref{table_2}.

\begin{table}[!t]
\caption{The accuracy of different machine learning methods for Rice dataset \cite{cinar2019}.}
\begin{center}
\begin{tabular}{|c|c|c|c|c|c|}
\hline
\textbf{Method Name}&\textbf{Accuracy}&\textbf{Recall}&\textbf{Specificity}&\textbf{Precision}&\textbf{F1-Score}\\
\hline
Logistic Regression (LR) & 93.02 & 92.26 & 93.58 & 91.35 & 91.80 \\
\hline
Support Vector Machine (SVM) & 92.83 & 91.70 & 93.68 & 91.53 & 91.62 \\
\hline
Decision Tree (DT) & 92.49 & 91.18 & 93.48 & 91.29 & 91.23 \\
\hline
Random Forest (RF) & 92.39 & 91.36 & 93.15 & 90.80 & 91.08 \\
\hline
Naive Bayes (NB) & 91.71 & 90.86 & 92.33 & 89.63 & 90.24 \\
\hline
k-Nearest Neighbor (k-NN) & 88.58 & 86.37 & 90.26 & 87.06 & 86.71 \\
\hline
Multi-layer Perceptron (MLP) & 92.86 & 92.17 & 93.36 & 91.04 & 91.60 \\
\hline
PSO & 92.71  & 90.92 & 94.04 & 91.94 & 91.43 \\
 & $\pm$ 0.16 & $\pm$ 0.27 & $\pm$ 0.26 & $\pm$ 0.32 & $\pm$ 0.19\\
\hline
Python PSO & 92.62 & 90.96 & 93.86 & 91.73 & 91.34 \\
 & $\pm$ 0.12 & $\pm$ 0.31 & $\pm$ 0.28 & $\pm$ 0.33 & $\pm$ 0.14 \\
\hline 
\end{tabular}
\label{table_1}
\end{center}
\end{table}

\begin{table}[!t]
\caption{The accuracy of different machine learning methods for Dry bean dataset \cite{koklu2020}.}
\begin{center}
\begin{tabular}{|c|c|c|c|c|c|}
\hline
\textbf{Method Name}&\textbf{Accuracy}&\textbf{Precision}&\textbf{Recall}&\textbf{Specificity}&\textbf{F1-score}\\
\hline
Support Vector Machine (SVM) & 93.13 & 94.45 & 94.03 & 98.77 & 94.23 \\
\hline
Decision Tree (DT) & 87.92 & 89.19 & 87.93 & 97.91 & 88.29 \\
\hline
k-Nearest Neighbor (k-NN) & 92.52 & 93.93 & 93.59 & 98.67 & 93.75 \\
\hline
Multi-layer Perceptron (MLP) & 91.73 & 93.11 & 92.68 & 98.53 & 92.88 \\
\hline
PSO & 91.50 & 92.47 & 92.75 & 98.51 & 92.61 \\
\hline
Python PSO & 91.76 & 92.70 & 93.01 & 98.56 & 92.85 \\
\hline 
\end{tabular}
\label{table_2}
\end{center}
\end{table}

%\subsection{Training}

%We trained the model from end to end to optimize both the retriever and the generative model together. Here, we aim to train a numerically literate model, which means that the model should be able to understand the numbers in the news and be able to perform mathematical operations on these numbers to arrive at a precise numerical value that will be used in the headline. We encapsulate the numbers in the headline of the news text with XML tags, as shown in the example in Table. 

\begin{figure*}[!ht]
\centering
\includegraphics[scale=0.341]{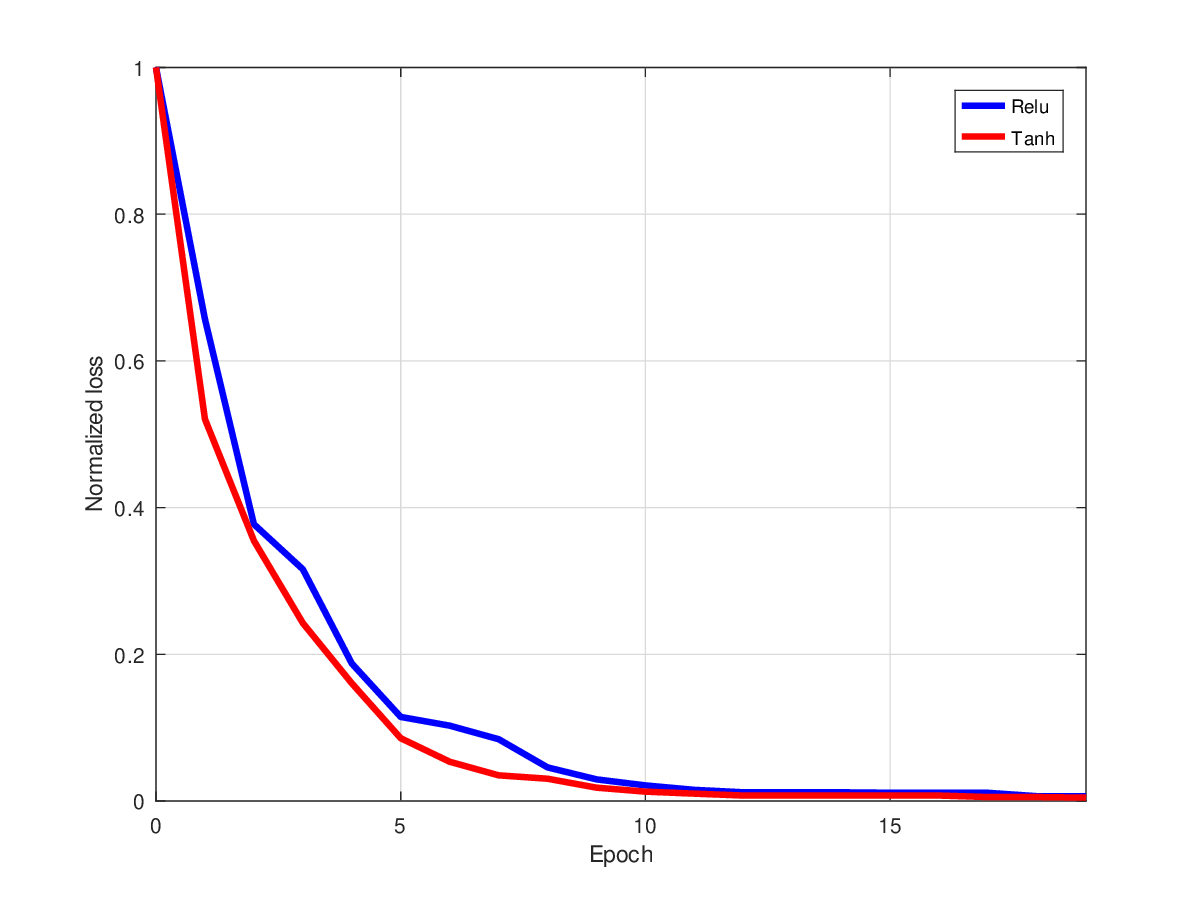}
\includegraphics[scale=0.341]{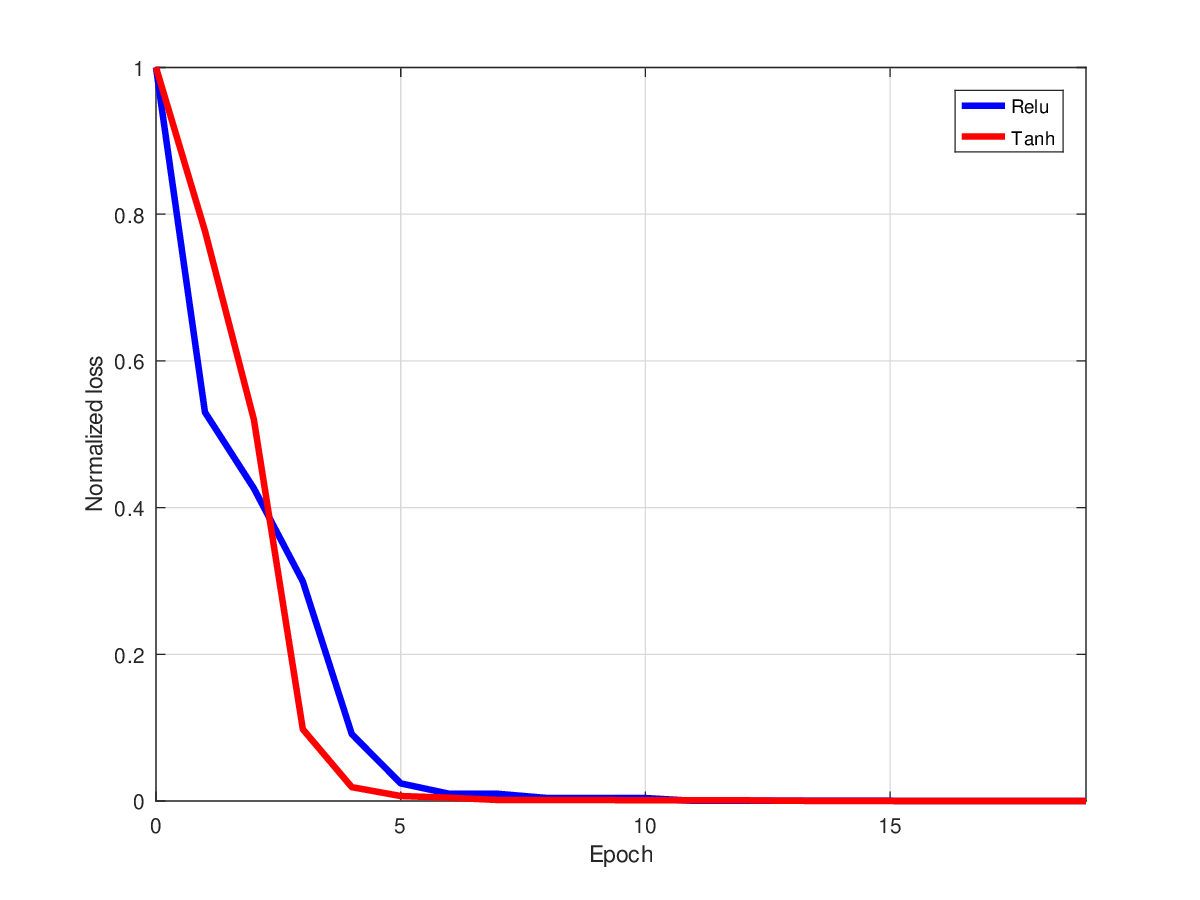}
\caption{The normalized loss function after each epoch is shown for Relu and Tanh functions in the proposed method using Octave. The left side of the plot is related to Figure \ref{fig:figure4} and the right side of the plot is related to Figure \ref{fig:figure5}.}
\label{fig:figure7}
\end{figure*}

\section{Discussion}

We have proposed a method that does weight updates without backpropagation. We compare the flow rate of loss value between the proposed method and the backpropagation-based MLP architecture. It is confirmed that the proposed method follows the backpropagation approach while improving the loss value. We tested our proposed method on two real datasets to validate the results obtained using MLP architecture. The last three rows of Tables \ref{table_1} and \ref{table_2} show the performance metrics close to each other. The neural network for the real datasets has more layers and nodes in each layer compared to the simulated datasets. The performance of the proposed method is similar to that of the backpropagation approach. We used a rectified linear unit (Relu) as a non-linear function for all the nodes while computing the loss value. We replaced a rectified linear unit (Relu) with tangent hyperbolic (Tanh) as a non-linear function. The loss values for two simulated datasets using Relu and Tanh functions are shown in Figure \ref{fig:figure7}. The underlying weight update operation is unaffected by changes in the non-linear functions present in the nodes.

One of the main advantages of the proposed method is individual computation by each node. The weight updates in each node are independent and can be performed parallelly without depending on the weight updates of other nodes. It resembles a neuronal firing by a neuron in a brain without much dependencies on the other neurons. In backpropagation, all the weights are updated simultaneously. Here, the weight update is independent of layers and nodes in the proposed method. In backpropagation, we update the weights when there is an error signal, and the weight updates are proportional to the error signal. In the proposed method, the weight updates are random without using an error signal. It may look like a random search and run away from the global optimum value for the network. However, the global best weight of a node restricts the runaway problem through a controlled measure of validation loss. We can construct arbitrary neural network architectures without following standard formats or processes. There may be an individual connection between a lower-layer node and a higher-layer node. The proposed method provides an opportunity to create the possibility of a complex neural network constructed with all sorts of connections between the nodes. The complex connectivity may lead to a close resemblance of brain neuronal activities. We need to study the limitations of arbitrary neural networks by exploring the proposed method.

One of the main limitations of the proposed method is the repeated computation of loss values, which increases with the nodes, layers, and data. It is sometimes redundant to compute the loss values repeatedly without any improvements in the loss values. There may be a functional mapping between the loss value and the weights of each node. If we can create the functional mapping between the loss value and the node weights, then the repeated computation may be reduced. 

There is no clear evidence of backpropagation in the brain when updating the synaptic weights. However, the representation formed in each layer resembles the trained weights of a neural network. Most neural networks are trained using a single global objective function, but biological neural networks rely on self-organization behavior \cite{schneider2024}. In our proposed method, each neuron is considered independent and aligns with the global objective. The neuron node receives the performance score and aligns itself in that direction. The neuron doesn't look at the other neuronal nodes for synaptic weight changes and keeps updating its weights. Thus, the weight update doesn't depend on the other nodes in the proposed method. The global best weights constrain and guide the neuron to reach optimal value. We tried to tie the global best weights as part of the correction in each particle within a node. This approach didn't yield fruitful results. Hence, we left individual particles within a node to act independently and use only the performance score for alignment. It is like tuning a filter by turning the knobs and observing the performance. We keep a reference while turning the knobs and choose the best performance. A similar concept is used in the particles to select the best weights. The objective loss function is split into many smaller functions. The smaller objective functions at each neuron may be related to the objective function and more concerned with choosing the best weight for the neuron. A collective behavior of nodes and particles shows the possibility of training a network in the proposed method.  

\section{Conclusion}

We proposed a method to update the node weights in a neural network. The proposed method overcomes the drawbacks of backpropagation and PSO. However, the proposed method has the drawback of redundant computation. We would like to reduce the computation by exploring the functional mapping between the nodes. In this paper, we explored MLP architecture using the proposed method on the simulated and real datasets. The performance metrics are similar to MLP architecture, which is trained using backpropagation. We would like to explore other neural network architectures like RNN, Transformers, and other networks to identify the limitations of the proposed method. We have open-sourced the code and data available at \href{https://github.com/dipkmr/train-nn-wobp/}{https://github.com/dipkmr/train-nn-wobp/}

%\section*{References}

%\bibliographystyle{acl_natbib}
\printbibliography

%%%%%%%%%%%%%%%%%%%%%%%%%%%%%%%%%%%%%%%%%%%%%%%%%%%%%%%%%%%%

\end{document}